\pdfoutput=1

\documentclass[11pt]{article}

\usepackage[final]{acl}

\usepackage{times}
\usepackage{latexsym}

\usepackage[T1]{fontenc}

\usepackage[utf8]{inputenc}

\usepackage{microtype}

\usepackage{inconsolata}

\usepackage{graphicx}
\usepackage{multirow}
\usepackage{multicol}
\usepackage{amssymb}
\usepackage{amsmath}
\title{A General Knowledge Injection Framework for ICD Coding}

\renewcommand{\thefootnote}{\fnsymbol{footnote}}

\author{
 \textbf{Xu Zhang}\textsuperscript{1,2} ,
 \textbf{Kun Zhang}\textsuperscript{1,2}\footnotemark[1],
 \textbf{Wenxin Ma}\textsuperscript{1,2} , \\
 \textbf{Rongsheng Wang}\textsuperscript{\textbf{1,2}} \textbf{,}
 \textbf{Chenxu Wu}\textsuperscript{\textbf{1,2}} \textbf{,}
 \textbf{Yingtai Li}\textsuperscript{\textbf{1,2}} \textbf{,}
 \textbf{S. Kevin Zhou}
 \textsuperscript{\textbf{1,2,3,4}}\footnotemark[1]\\
 \textsuperscript{1} School of Biomedical Engineering, Division of Life Sciences and Medicine, USTC\\
 \textsuperscript{2} MIRACLE Center, Suzhou Institute for Advance Research, USTC
\\
 \textsuperscript{3}Jiangsu Provincial Key Laboratory of Multimodal Digital Twin Technology
\\
 \textsuperscript{4}State Key Laboratory of Precision and Intelligent Chemistry, USTC
\\
{\tt\small xu\_zhang@mail.ustc.edu.cn kkzhang@ustc.edu.cn skevinzhou@ustc.edu.cn}
}

\begin{document}
\maketitle

\footnotetext[1]{Corresponding authors}

\renewcommand{\thefootnote}{\arabic{footnote}}
\setcounter{footnote}{0}

\begin{abstract}
ICD Coding aims to assign a wide range of medical codes to a medical text document, which is a popular and challenging task in the healthcare domain. To alleviate the problems of long-tail distribution and the lack of annotations of code-specific evidence, many previous works have proposed incorporating code knowledge to improve coding performance. However, existing methods often focus on a single type of knowledge and design specialized modules that are complex and incompatible with each other, thereby limiting their scalability and effectiveness. To address this issue, we propose \textbf{GKI-ICD}, a novel, general knowledge injection framework that integrates three key types of knowledge, namely ICD Description, ICD Synonym, and ICD Hierarchy, without specialized design of additional modules. The comprehensive utilization of the above knowledge, which exhibits both differences and complementarity, can effectively enhance the ICD coding performance. Extensive experiments on existing popular ICD coding benchmarks demonstrate the effectiveness of GKI-ICD, which achieves the state-of-the-art performance on most evaluation metrics. Code is available at \url{https://github.com/xuzhang0112/GKI-ICD}.
\end{abstract}


\section{Introduction}

\begin{figure}[t]
    \centering
  \includegraphics[width=\linewidth]{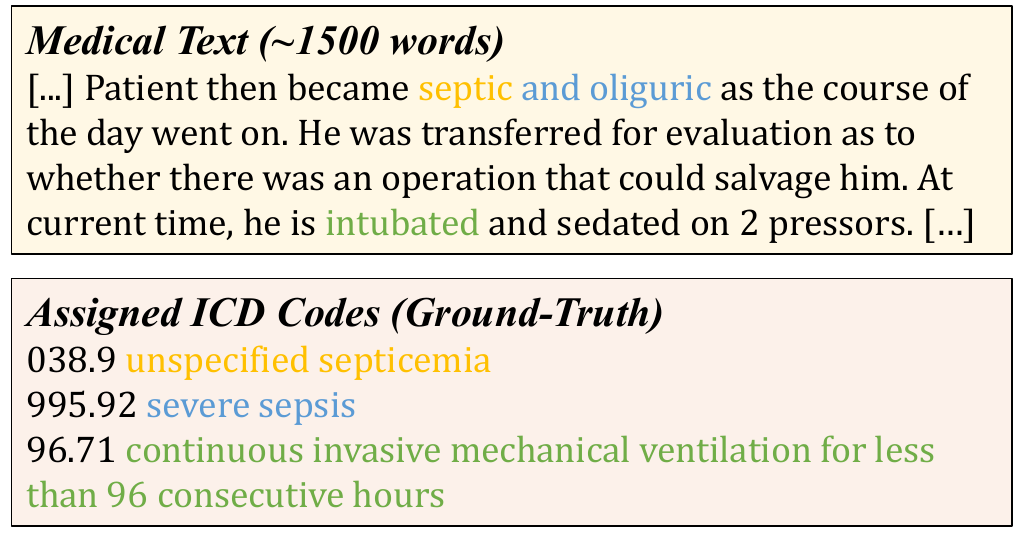}
  \caption{{An example of ICD coding}: Occurrence of multiple codes and noisy content in a long medical text document makes it hard to link each ICD code to its corresponding evidence (marked in same color), explaining the necessity of incorporating code-specific knowledge.}
  \label{fig:moti}
\end{figure}

International Classification of Diseases (ICD) \footnote{https://www.who.int/standards/classifications/classification-of-diseases} is a globally used medical classification system, developed by the World Health Organization to classify diseases, symptoms, procedures, and external causes.
The ICD coding task aims to assign the most accurate ICD codes to clinical texts, typically discharge summaries, for further medical billing and clinical research. 
Two main challenges arise in the ICD coding process~\cite{survey_replication}. First, there is a tremendous number of ICD codes to assign in clinical practice, whose distribution is extremely long-tailed, and most of which are lacking in enough training samples. 
Second, as shown in Figure \ref{fig:moti}, the occurrence of multiple ICD codes within a long medical document makes it hard for models to accurately link each ICD code with its corresponding evidence fragments. Human coders do not annotate the evidence of the ICD codes assigned by them, due to the complexity of this operation, only leaving document-level annotations to each medical document.


In recent years, numerous studies~\cite{survey_method} have explored the incorporation of ICD code-related knowledge to assist models in precisely locating evidence fragments related to specific ICD codes, thereby effectively and efficiently improving coding performance.
Generally, three types of knowledge are involved in ICD coding:\emph{ ICD Description, ICD Synonym, and ICD Hierarchy}. Specifically, 1) ICD Description refers to the meaning of each ICD code, which is directly related to the coding process. Language models can leverage semantic mapping to identify the most relevant evidence fragments within a long medical text document, facilitating accurate classification. 2) ICD Synonym addresses the diversity of medical terminology, as a single ICD code may have multiple linguistic expressions. Incorporating synonyms helps the model recognize different variants of the same code, enhancing its robustness. 3) ICD Hierarchy organizes the relationships between codes. With tens of thousands of codes in ICD-9, these codes are not entirely independent. ICD Hierarchy provides a structured relationship between codes, particularly grouping rare codes with more common ones. 
\emph{Inherently, these three types of knowledge exhibit both differences and complementarity.}



However, existing methods typically focus on only one of these different types of knowledge and design \textbf{specialized network architectures} accordingly, making it hard to integrate other complementary knowledge. To utilize synonym knowledge, current approaches often employ a multi-synonym-attention mechanism, where each query corresponds to a synonym~\cite{msmn,msam}. To incorporate hierarchical knowledge, methods primarily rely on graph neural networks, treating the hierarchical structure as an adjacency matrix to aggregate code representations~\cite{msatt-kg,dkec}. Since these methods design specialized modules for individual knowledge types, the complexity of these modules makes it difficult to scale to advanced models. More importantly, the incompatibility between these specialized modules hinders their integration into a unified model, preventing the comprehensive utilization of all knowledge types.





To address the above issue, we propose GKI-ICD, a novel synthesis-based multi-task learning fr to inject knowledge. In contrast to existing methods that often struggle with complex architectural designs and integration challenges, our method jointly leverages all types of knowledge without relying on specialized modules. Specifically, GKI-ICD consists of two key components: guideline synthesis and multi-task learning. The guideline synthesis incorporates ICD code knowledge to synthesize a guideline, ensuring that all the knowledge relevant to the raw sample is embedded within the guideline. Meanwhile, the multi-task learning mechanism requires the model to not only correctly classify the original samples but also make accurate predictions based on the synthesized guidelines. Additionally, it encourages the model to align the information extracted from the raw samples with that from the provided guidelines as closely as possible, thereby facilitating effective knowledge integration. 


Our main contributions are summarized as:
\begin{itemize}
\item[$\bullet$] To our knowledge, we are the first to inject ICD code knowledge \textbf{without requiring any additional specially-designed networks or prompts}, thus being able to integrate the three kinds of ICD code knowledge separately utilized before.

\item[$\bullet$] We propose a novel synthesis-based multi-task learning mechanism, including guideline synthesis and multi-task learning, to inject ICD code knowldge into the coding model.

\item[$\bullet$] We achieve state-of-the-art performance on most evaluation metrics on the ICD coding benchmarks MIMIC-III and MIMIC-III-50, proving not only the effectiveness of our knowledge injection framework, but also the necessity of multiple knowledge integration. 

\end{itemize}

\section{Related Work}

\subsection{ICD Coding Network}

The automatic ICD coding task is well established in the healthcare domain, and most of the approaches fisrt encode the discharge summary with a text encoder, and then use a label attention mechanism to attend, aggregate, and make predictions.

\textbf{Text encoder.} Early ICD coding methods~\cite{caml,laat,mrcnn,effectivecan} primarily utilized convolutional neural networks (CNNs), recurrent neural networks (RNNs), and their variants as backbones, while recent approaches~\cite{plmicd,plmca} have been based on pretrained language models (LMs). Besides, large language models (LLMs) have been proved to perform worse on this task~\cite{llmicd}, compared to fine-tuned small models.

\textbf{Label attention.} Instead of making predictions based on a pooled vector, label attention use a linear layer to compute relationships between each ICD code and each token in the clinical text, aggregate different information for different codes and then make predictions~\cite{caml}. Subsequently, this linear layer was replaced by a multilayer perceptron~\cite{laat}, and was finally replaced by the standard cross attention~\cite{plmca}, both improving the training stability and slightly enhancing its performance.
  
\subsection{Knowledge Injection}

Considering the rich prior knowledge in biomedical domain, many efforts have been made to incorporate medical knowledge to enhance model performance on ICD coding tasks. Knowledge injection methods can generally be divided into two categories: task-agnostic and task-specific. 

\textbf{Task-agnostic knowledge.} Extensive biomedical corpora, such as electronic health records and biomedical academic papers, can be utilized for pretraining language models. These pretrained models, including BioBERT~\cite{biobert}, ClinicalBERT~\cite{clinicalbert}, PubMedBERT ~\cite{pubmedbert} and RoBERTa-PM~\cite{roberta}, serve as powerful biomedical text encoders, significantly enhancing the performance of downstream tasks, including ICD coding.

\textbf{Task-specific knowledge.} Task-specific knowledge refers to information related to ICD codes, such as the meaning of each code and the hierarchical structure among codes. By injecting this kind of knowledge during the fine-tuning stage, the model's performance on the ICD coding task can be improved. MSATT-KG~\cite{msatt-kg} leverages graph convolutional neural network to capture the hierarchical relationships among medical codes and the semantics of each code. ISD~\cite{isd} proposes a self-distillation learning mechanism, utilizing code descriptions help the model ignore the noisy text in clinical notes. MSMN~\cite{msmn} uses multiple synonyms of code descriptions to initialize the code query embeddings. KEPTLongformer~\cite{keptlongformer} incorporates a medical knowledge graph for self-alignment contrastive learning, and then adds a sequence of ICD code descriptions as prompts in addition to each clinical note as model input. DKEC~\cite{dkec} propose a heterogeneous graph network to encode knowledge from multiple sources, and generate knowledge-based queries for each ICD code. MRR~\cite{mrr} and AKIL~\cite{akil} incorporates diagnosis-related group (DRG) codes, current procedural terminology (CPT) codes, and medications prescribed to patients to generate a dynamic label mask, which can help down-sample the negative labels and focus the classifier on candidate labels.
Unlike previous methods that design specialized networks for knowledge injection, we propose a general knowledge injection framework, making it applicable to various models and diverse types of knowledge.

\section{Methodology}


We first provide an overview in Section \ref{task}, highlighting the key differences between our proposed GKI-ICD and previous works. Next, we elaborate on its details in Section \ref{framework}. In addition, we briefly describe the ICD coding network adopted in our work in Section \ref{arch}.

\subsection{Overview}
\label{task}

Typically, the ICD coding task involves optimizing an ICD coding network to assign specific ICD codes to the given medical text, defined as:
\begin{equation}
\theta^* = \arg\min_{\theta} \mathcal{L}(f(\mathbf{x}; \theta),y),
\end{equation}
where $\mathbf{x}$ represents the input medical text and $\mathbf{y}$ denotes the corresponding ground-truth ICD codes, $\theta$ denotes the parameters of the ICD coding network.

To further boost performance, existing methods \cite{msatt-kg,keptlongformer,msmn,dkec,msam,corelation} generally devise additional neural networks to inject knowledge, i.e., 
\begin{equation}
\theta^*;\theta_i^* = \arg\min_{\theta;\theta_i} \mathcal{L}(g_i(\mathbf{x}; \theta;\theta_i),y),
\end{equation}
where $g_i$ is a neural network specially designed to incorporate the $i$-th type of knowledge, and $\theta_i$ denotes the corresponding additional module parameters. To be specific, $g_i$ can be graph neural networks for hierarchy knowledge \cite{msatt-kg} or multi-synonym attention networks for synonym knowledge \cite{msmn}, etc.

\begin{figure*}[htbp]
  \includegraphics[width=\linewidth]{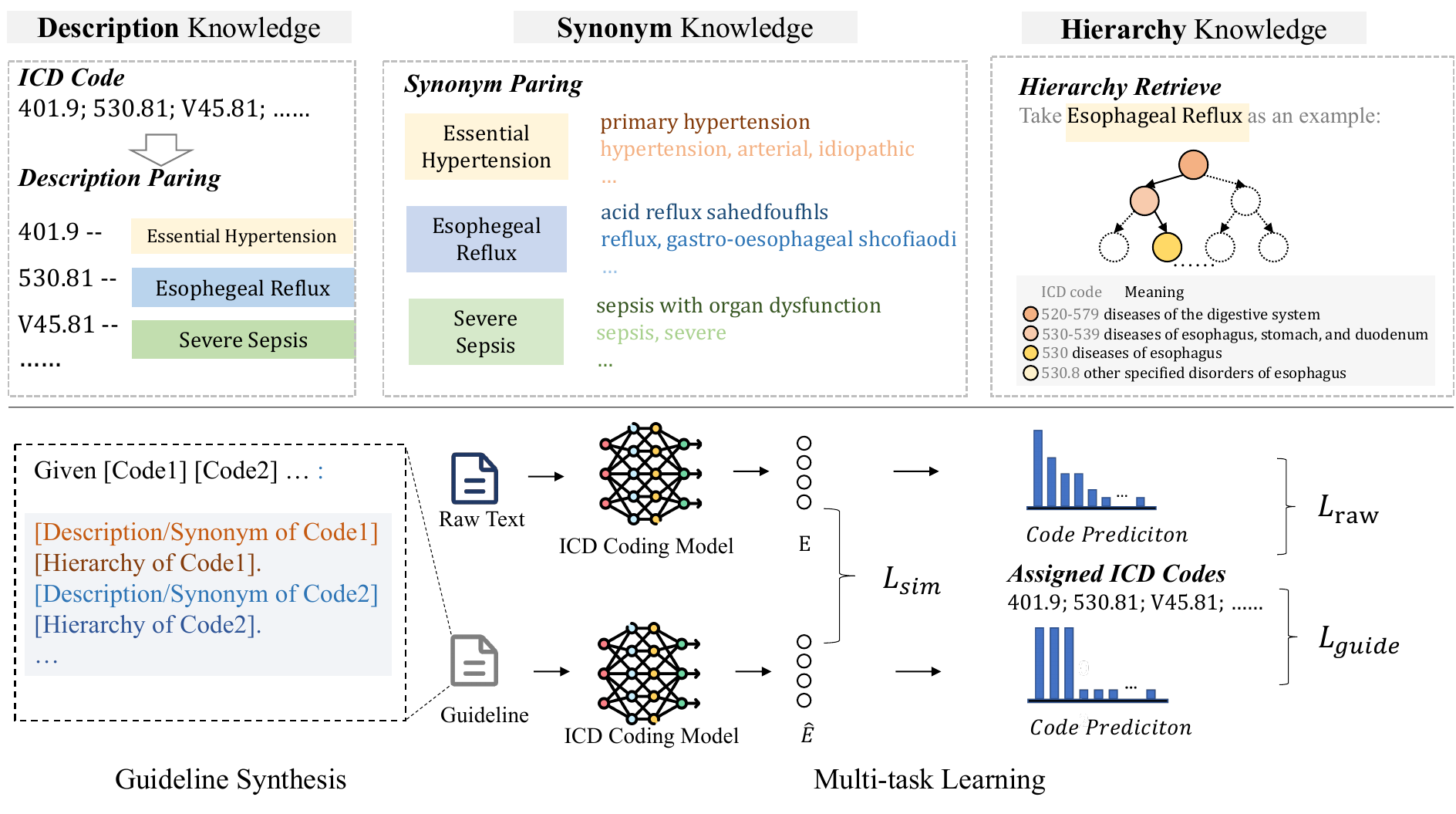}
  \caption{Our proposed general knowledge injection training framework for ICD coding, GKI-ICD. For each training sample, we first retrieve code-specific knowledge to synthesize a guideline, and then use this guideline and multi-task learning to inject knowledge into the model. Note that our method only incorporates knowledge in the training stage, which has no effect on the computation cost of the model during the inference stage.}
  \label{fig:method}
\end{figure*}

However, considering these extra modules are complex and hard to integrate simultaneously, our approach aims to propose a new training framework that can inject knowledge without extra parameters. By leveraging knowledge to synthesize guidelines $\hat{x}$ and modifying the training pipeline, we enable the injection of all necessary knowledge to be free of extra parameters or interactions. The proposed knowledge injection framework can be defined as:
\begin{equation}
\theta^* = \arg\min_{\theta} \mathcal{L'}(f(\mathbf{x}; \mathbf{\hat{x}}; \theta),y),
\end{equation}
where $f$ is the simplest ICD coding network, having the merit to be adapted to any state-of-the-art network. In the following, as illustrated in Fig.~\ref{fig:method}, we give the details including guideline synthesis based on knowledge in \ref{synthesis} and multi-task learning based on synthetic guidelines in \ref{training}.

\subsection{Proposed Method}
\label{framework}

\subsubsection{Guideline Synthesis}
\label{synthesis}
Given a medical text document with a set of ICD codes, we synthesize a guideline by retrieving relevant knowledge associated with each ICD code assigned to this document. This guideline can assist the model in learning to localize evidence fragments and make accurate code predictions. 



\textbf{Description parsing.}
Given document-level annotations $\{y_i\},  y_i\in \{0, 1\}$, we can extract the set of ICD codes present in the document, referred to as the positive code set. Let the full code set be denoted as $\{c_1, ..., c_n\}$, and the positive code set be represented as:
\begin{equation}
    C_p=\{c_i|y_i=1\}, 
\end{equation}
Since each code $c_i$ has an offical description in ICD-9, it can be denoted as $\text{Description}(c_i)$. 
We can easily retrieve the descriptions of these assigned ICD codes in the positive code set:
\begin{equation}
    D_p = \{\text{Description}(c_i)|c_i \in C_p\},
\end{equation}
which can be used to build the synthetic guideline. We remove the term "NOS" (Not Otherwise Specified) to standardize expressions.


\textbf{Synonym replacement.} To enhance the diversity of synthetic samples and enrich the representation of each code, we incorporate synonyms~\cite{msmn} derived from biomedical knowledge bases. For instance, code 401.9 in ICD-9 is defined as "unspecified essential hypertension", but may be referred to in alternative terminologies such as "primary hypertension" or "hypertension nos." These variations can be systematically identified within the Unified Medical Language System (UMLS) ~\cite{umls}, a structured repository of biomedical terminologies that provides multiple synonymous expressions for all ICD codes.


We first map each ICD code to its corresponding Concept Unique Identifier (CUI) and extract the English synonyms associated with the same CUI. 
For a specific code $c_i$ with multiple synonyms, we randomly sample one of these synonyms, i.e.,
\begin{equation}
    s_i=\text{Synonym}(c_i)\sim\{s_i^1,s_i^2,...,s_i^k\},
\end{equation}
where $s_i^k$ is the $k$-th synonym.
Then we replace the code descriptions with these sampled synonyms to obtain:
\begin{equation}
    S_p = \{s_i|c_i \in C_p\}.
\end{equation}
This synonym substitution strategy facilitates diverse and robust code representation and enhances the adaptability to real-world medical texts.

\textbf{Hierarchy retrieve.}
Another important source of prior knowledge is the hierarchical relationships between ICD codes. For example, code 038.9 ("unspecified septicemia") belongs to code groups 030-041 ("other bacterial diseases") and 001-139 ("infectious and parasitic diseases"), which include many similar but distinct codes. The hierarchical information of a code can be defined as $\text{Hierarchy}(c_i)$, which contains the descriptions of all the groups to which this ICD code belongs.


While code hierarchy knowledge is commonly incorporated by designing graph neural networks with predefined adjacency matrices, we assume that the language model can adaptively retrieve semantic information from the complete descriptions. This is achieved by simply adding all hierarchical knowledge to the guideline as:
\begin{equation}
    H_p = \{\text{Hierarchy}(c_i)|c_i\in C_p\}.
\end{equation}



\textbf{Shuffle and concatenate}. Next, we shuffle the order of the assigned codes $C_p$, replace them with their descriptions and hierarchical descriptions, and concatenate them to form a long string sequence $\hat{x}$.

Thus, for each training sample $(x, y)$, we generate a synthetic guideline $\hat{x}$ that encapsulates the relevant knowledge of the ICD codes assigned to the raw training sample.

\subsubsection{Multi-task Learning}
\label{training}

\textbf{Retrieve and prediction from raw text}. In an ordinary setting, the ICD coding model makes a binary prediction based on the raw clinical document as:
\begin{align}
    L_{raw} = L_{BCE}(f(x),y),
\end{align}
where $x$ is the medical document, and $y$ is the binary vector whose dimension equals the total number of ICD codes. The predictions are supervised by the binary labels using cross-entropy loss:
\begin{align}
  L_{BCE} = -\frac{1}{C} \sum_{i=1}^{C}
( y_i \log p_i + (1-y_i) \log (1-p_i)),
\end{align}
where $C$ is the total number of ICD-9 codes and $i$ refers to the dimension of the predicted vector and ground truth vector.

\textbf{Retrieve and prediction from guideline}. Given that the guideline encapsulates all the semantic information of the assigned ICD codes, the model is guided to retrieve code-specific details and predict the corresponding ICD codes. We employ
\begin{align}
    L_{guide} = L_{BCE}(f(\hat{x}),y),
\end{align}
to achieve this goal.
This guideline, free from noisy content such as social and family history, simplifies the assignment of ICD codes and facilitates smoother learning for the ICD coding model.

\textbf{Semantic similarity constraint}: We apply a similarity loss function to enforce consistency between the code-specific representations aggregated from the raw sample and its corresponding guideline. Only the assigned ICD codes are considered, using the binary ground truth vector to select the aggregated vector of these positive codes by:
\begin{align}
    E &= y \odot E \\
    \hat{E}&=y \odot \hat{E},
\end{align}
where $E \in R^{C\times D}$ and $\hat{E} \in R^{C\times D}$ are code-specific representations obtained by the ICD coding model elaborated in Section \ref{arch}.
Then, we compute the similarity between each of the two retrieved features: one from a normal clinical document, and the other from the guideline, as the loss function:
\begin{align}
    L_{sim}&=1-cosine(E,\hat{E}),
\end{align}
to make them consistent in the semantic space.

The total optimization function can be formulated as:
\begin{align}
    L=L_{raw}(x,y)+L_{guide}(\hat{x},y)+\lambda L_{sim}(E,\hat{E}),
\end{align}
where $\lambda$ is a coefficient to control the similarity, considering the gap between theoretical code knowledge and clinical code expressions.

\subsection{Model Architecture}
\label{arch}

\begin{figure}[h]
    \centering
    \includegraphics[width=\linewidth]{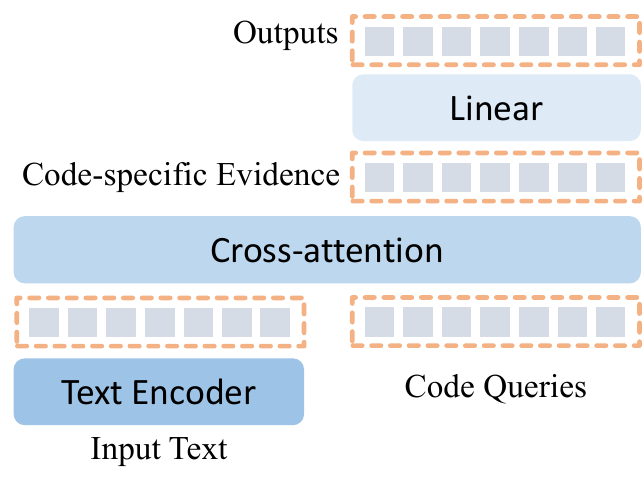}
    \caption{The model architecture adopted in our work.}
    \label{fig:model}
\end{figure}

Following PLM-CA~\cite{plmca}, 
our model comprises an encoder and a decoder. The encoder transforms a sequence of $N$ tokens into a sequence of contextualized token representations $H \in \mathbb{R}^{N \times D}$. We use RoBERTa-PM~\cite{roberta}, a transformer pre-trained on PubMed articles and clinical notes, as the encoder. 
However, the length of clinical documents is larger than the max input length of RoBERTa-PM, so we chunk the raw document text into pieces, feed them into the PLM separately, and concatenate them along with the axis of length in feature space. For simplicity, we describe this process as:
\begin{align}
    \mathbf{H} &= \text{PLM}(x).
\end{align}

After obtaining the contextual representations of the input text, we use a standard cross attention to aggregate information for different ICD codes. The code-specific evidence $E_i$ can be obtained by:
\begin{align}
    A_i &= \text{softmax}(Q_i (HW_k))^T),\\
    E_i &= \text{layernorm}(A_i (HW_v)),
\end{align}
where $Q_i \in R^D$ is the learnable code query of the $i$-th ICD code, $A_i \in \mathbb{R}^{C \times N}$ is the attention matrix from the $i$-th code to the input text, and $W_k$, $W_v \in \mathbb{R}^{D \times D}$ are the linear transform matrices.

Based on the aggregated evidence of the $i$-th ICD code, a linear classifier is applied to compute the predicted probability for $i$-th ICD code:
\begin{equation}
    \hat{y}_i = \text{sigmoid}(E_iW_i),
\end{equation}
where $W_i \in \mathbb{R}^{D}$ is an independent linear classifier applied to the $i$-th ICD code.

\section{Experiments} \label{Experiments}
\subsection{Experiment Setting}

\begin{table}[h]
\centering
\begin{tabular}{lrrrr}
\hline
Dataset & $N_{Train}$ & $N_{Dev}$ & $N_{Test}$ & $N_{Codes}$ \\
\hline
Full & 47,723 & 1,631 & 3,372 & 8,929 \\
Top-50 & 8,066 & 1,573 & 1,729 & 50 \\
\hline
\end{tabular}
\caption{Statistics of MIMIC-III Dataset Splits. $N_{Train}$, $N_{Dev}$ and $N_{Test}$ refer to the number of samples in the train, development and test split. $N_{Codes}$ refers to the number of unique ICD codes in the whole dataset.}
\label{tab:dataset_stats}
\end{table}

\textbf{Dataset.} We use the MIMIC-III dataset ~\cite{mimic}, which is the largest publicly available clinical dataset. We follow the
experimental setting of ~\citet{caml} to
form MIMIC-III-Full and MIMIC-III-Top-50. The statistical data for the two datasets are presented in Table \ref{tab:dataset_stats}. Following the setting of ~\citet{plmca}, we train and test the models on raw text, only truncating all documents to a maximum of 8,192 tokens without any other pre-processing.

\textbf{Evaluation metrics.} Following previous work ~\cite{caml}, we evaluate our method using both macro and micro F1 and AUC metrics, mean average precision (MAP), and precision at K (P@K) that indicates the proportion of the correctly predicted labels in the top-K predictions. For MIMIC-III-Full Dataset, we set K as 8, 15, while for the MIMIC-III-Top-50 Dataset, we set K as 5.

\textbf{Implementation details.}
We implement our model in PyTorch ~\cite{pytorch} on a single NVIDIA H20 96G GPU. We use the Adam optimizer and the learning rate is initialized to 5e-5. We train the model for 12 epochs, the learning rate increases in the first 2000 steps, and then decays linearly in the further steps. The batch size is 8, which indicates that there are 8 raw samples and 8 synthetic guidelines in a batch in our proposed framework. We initialize each code query with its ICD description by encoding the text and employing a maximum pooling, inspired by \citet{label-embedding}. We use R-Drop~\cite{rdrop} regularization techniques to alleviate overfitting, and set $\alpha$ as 5 for MIMIC-III-Full Dataset and 10 for MIMIC-III-Top-50 Dataset.

\subsection{Comparison with SOTA models}

\begin{table*}[t]
\resizebox{\textwidth}{!}{
\begin{tabular}{l cccccc  ccccc}
\hline
\multirow{3}{*}{Models} & \multicolumn{6}{c}{MIMIC-III-Full} & \multicolumn{5}{c}{MIMIC-III-Top-50} \\ 
\cline{2-12} 
& \multicolumn{2}{c}{AUC} & \multicolumn{2}{c}{F1} & \multicolumn{2}{c}{P@K} & \multicolumn{2}{c}{AUC} & \multicolumn{2}{c}{F1} & \multirow{2}{*}{P@5} \\
\cline{2-11} 
& Macro      & Micro      & Macro      & Micro     & P@8        & P@15       & Macro      & Micro      & Macro      & Micro     &                      \\
\hline

CAML ~\cite{caml}  & 0.895      & 0.986      & 0.088      & 0.539     & 0.709      & 0.561      & 0.875      & 0.909      & 0.532      & 0.614     & 0.609           \\
MSATT-KG ~\cite{msatt-kg}             & 0.910      & 0.992      & 0.090      & 0.553     & 0.728      & 0.581      & 0.914      & 0.936      & 0.638      & 0.684     & 0.644             \\
MSMN ~\cite{msmn}                   & 0.950      & 0.992      & 0.103      & 0.584     & 0.752      & 0.599      & 0.928      & 0.947      & 0.683      & 0.725     & 0.680  \\      
KEPTLongformer ~\cite{keptlongformer}
& -          & -          & 0.118      & 0.599     & 0.771      & 0.615      & 0.926      & 0.947      & 0.689     & 0.728     & 0.672   \\
PLM-ICD~\cite{plmicd}& 0.926& 0.989 & 0.104 & 0.598 & 0.771 & 0.613&
0.910&0.934&0.663&0.719&0.660\\
PLM-CA ~\cite{plmca}& 0.916 & 0.989    &   0.103    &  0.599 & 0.772 & 0.616  & 0.916 & 0.936 & 0.671 & 0.710 & 0.664   \\
CoRelation ~\cite{corelation}& 0.952 & 0.992    &   0.102    &  0.591 & 0.762 & 0.607  & 0.933 & 0.951 & \textbf{0.693} & 0.731 & \textbf{0.683}   \\
GKI-ICD (Ours) & \textbf{0.962} & \textbf{0.993} & \textbf{0.123} & \textbf{0.612} & \textbf{0.777} & \textbf{0.624} 
 & \textbf{0.933}	& \textbf{0.952}&	0.692 &	\textbf{0.735}&	0.681
\\

\hline      

 MRR~\cite{mrr} &0.949 & 0.995 & 0.114& 0.603 &0.775 &0.623& 0.927 &0.947 &0.687& 0.732& 0.685 \\
 
 AKIL~\cite{akil}&0.948& 0.994 & 0.112& 0.605& 0.784 & 0.637 & 0.928& 0.950 & 0.692& 0.734& 0.683\\
\hline
\end{tabular}}
\caption{Comparison with previous SOTA methods. Note that MRR and AKIL rely on DRG codes, CPT codes and medications, which are additionally annotated to each sample by human coders. We list these methods for reference although directly comparing our method with them is unfair.}
\label{table:compare}
\end{table*}

To demonstrate the superiority of proposed GKI-ICD framework,
we compare it with the state-of-the-art methods for ICD coding.

\textbf{Methods without knowledge.} CAML~\cite{caml} is a CNN-based model, which is the first work to propose explainable ICD coding; PLM-ICD~\cite{plmicd} and PLM-CA~\cite{plmca} are transformer-based models, which are popular these years in ICD coding.

\textbf{Methods with extra knowledge.} MSATT-KG~\cite{msatt-kg} captures code hierarchical relationships with graph neural networks; MSMN~\cite{msmn} proposes multi-synonym-attention to learn diverse code representations; KEPTLongformer~\cite{keptlongformer} adds the description of each ICD code to a long prompt; CoRelation~\cite{corelation} integrates context, synonyms and code relationships to enhance the learning of ICD code representations.

\textbf{Methods with additional human annotated data.}
 AKIL~\cite{akil} and MRR~\cite{mrr} improve ICD coding using additional human annotations, e.g., DRG codes, and CPT codes. Although directly comparing our methods with them is unfair, we list them for reference.

\textbf{Methods using LLMs.}
Currently, LLMs under zero-shot prompting perform worse than fine-tuned PLMs on ICD coding tasks, according to \citet{llmicd}.
To our knowledge, no published work has applied fine-tuned LLMs to ICD coding to achieve comparable performance to PLMs. 

Table \ref{table:compare} shows the quantitative results of these approaches on MIMIC-III-Full and MIMIC-III-Top-50. Our method outperforms state-of-the-arts significantly on all evaluation metrics. Specifically, compared with PLM-CA, on whose basis our model builds, our method obtains
4.6\% improvement on MacroAUC and 2.0\% improvement on MicroAUC, respectively, on MIMIC-III-Full. It also obtains 1.7\% gains on Macro AUC and 2.6\% gains on Micro AUC on MIMIC-III-Top-50, which only considers the most common ICD codes in MIMIC-III-Full.
Moreover, even compared with methods that rely on extra annotated inputs, e.g., AKIL and MRR, our method shows comparable performance and is even better on many metrics.
The improvement shows the effectiveness of GKI-ICD for using knowledge-based synthetic data to guide the learning process, and further verifies that jointly using the real samples and synthetic samples can obtain more accuracy.

\begin{table}[h]
\resizebox{\linewidth}{!}{
\begin{tabular}{l ccccc}
\hline
\multirow{2}{*}{Models} & \multicolumn{2}{c}{AUC} & \multicolumn{2}{c}{F1}& \multirow{2}{*}{MAP} \\
\cline{2-5}
& Macro      & Micro      & Macro      & Micro   &  \\
\hline
w/o knowledge & 0.917 & 0.989    &   0.109    &  0.606 &0.653  \\
\hline
w/ desc &   0.960    &   0.993    &    0.118    &  0.609&0.658  \\ 
w/ desc + syn & 0.962 & 0.993 & 0.123 & 0.611 &0.660\\
w/ desc + hie & 0.962 & 0.993 &  0.123 & 0.611&0.661\\ 
w/ desc + syn + hie & \textbf{0.962}&\textbf{0.993}&\textbf{0.123}&\textbf{0.612}&\textbf{0.661}\\
\hline
\end{tabular}}
\caption{Ablation of multiple knowledge injection on MIMIC-III-Full Dataset. The abbreviations "desc", "syn", "hie" stand for description knowledge, synonym knowledge and hierarchy knowledge, respectively. We apply our proposed knowledge injection training framework to the baseline model, and add different types of ICD code knowledge. Different from PLM-CA, all these models use R-drop regularization techniques and truncate input text into 8192 tokens, not 6144 tokens. }
\label{table:multiple_knowledge}
\end{table}

\begin{figure*}[t]
    \centering
    \includegraphics[width=\linewidth]{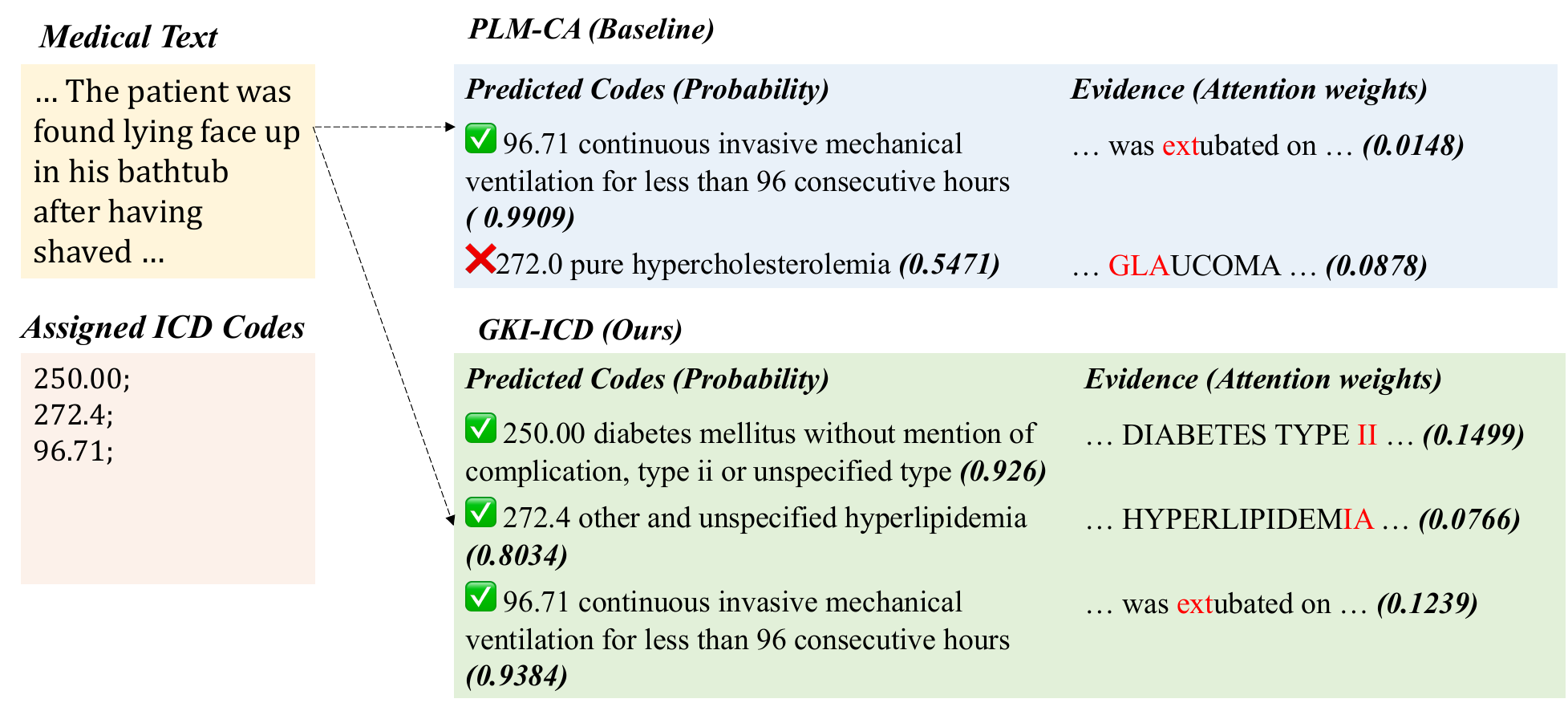}
    \vspace{-5mm}
    \caption{Case Study on MIMIC-III-Top-50 Dataset. We visualize the predicted ICD codes and the retrieved evidence of PLM-CA and our method. The red means the token which gains the greatest attention weight.}
    \label{fig:case}
    \vspace{-3mm}
\end{figure*}

\subsection{Ablation Study}
We conduct extensive ablation studies on MIMIC-III-Full dataset to verify the effectiveness of each component of our method.

\textbf{Effectiveness of proposed knowledge injection training framework}.
To address the challenges of long-tailed distribution and missing annotations, GKI-ICD injects knowledge through synthetic sample generation and multi-task learning. As shown in Table \ref{table:multiple_knowledge}, after incorporating any type of ICD code knowledge, the model demonstrates improvements across various evaluation metrics, highlighting the effectiveness of GKI-ICD and the importance of knowledge infusion. Furthermore, the model's performance is further enhanced by integrating all kinds of knowledge, demonstrating the compatibility of our approach with diverse types of knowledge and its potential for broader applications.

\textbf{Effectiveness of integrating multiple types of ICD code knowledge}. The impact of integrating multiple types of ICD code knowledge is explored. In addition to ICD code definitions, we incorporate synonym knowledge from a medical knowledge graph and hierarchy knowledge defined in ICD-9 system. These additional knowledge sources can be seamlessly integrated into GKI-ICD framework as supplementary information. As shown in Table \ref{table:multiple_knowledge}, incorporating richer knowledge enhances the ICD coding performance. This improvement highlights the importance of leveraging diverse and structured medical knowledge to better capture the semantic and relational nuances of ICD codes, leading to more accurate and robust predictions.




\begin{table}[t]\tiny
\centering
\resizebox{\linewidth}{!}{
\begin{tabular}{l cc}
\hline
Code Frequency & PLM-CA & GKI-ICD \\
\hline
>500 & 0.684 & 0.687\\
101-500 &0.508 &0.509 \\
51-100 & 0.413& 0.420\\
11-50 & 0.293& 0.322\\
1-10 & 0.029& 0.132\\
\hline
\end{tabular}
}
\caption{Comparison of F1-scores of PLM-CA and GKI-ICD on different code groups on MIMIC-III-Full Dataset.}
\label{table:freq}
\vspace{-5mm}
\end{table}

\subsection{Effectiveness on Rare Codes}

We classify the ICD codes into groups based on their frequencies in the training set, and test the F1 scores on different groups separately. As shown in Table \ref{table:freq}, GKI-ICD leads to improved accuracy across all code groups. Specifically, for rare codes (occurrence <= 10), GKI-ICD demonstrates an improvement of 0.103 micro-F1 score over PLM-CA, highlighting its superior capability in handling rare codes, as well as its potential to address other long-tailed distribution problem.

\subsection{Case Study}
We visualize an example from the test set, as shown in Figure \ref{fig:case}, comparing the attention weights and predictions before and after knowledge injection. Before knowledge injection, only half of the codes are correctly predicted by the model, and the evidence of the false positive code "272.0" is totally irrelevant to this code. However, after knowledge injection, the predicted codes are the same as the ground truth. Notably, the model pays attention to "Diabetes Type II", which is specially mentioned in the description of code "250.00". Moreover, the model pays more attention to the word "extubation", which is related to code "96.71", compared to the baseline. These changes substantiate the efficacy of knowledge injection.

\section{Conclusion}

In this paper, we propose GKI-ICD, a novel, general knowledge injection framework, which integrates multiple kinds of ICD code knowledge for guideline synthesis and inject code knowledge to the ICD coding model via multi-task learning.
Experimental results demonstrate that our proposed method outperforms the baseline models and is even comparable to models relying on extra human annotations. 
In addition, our framework does not make any changes to model architecture, thus being easy to be applied to other multi-label classification problems, using label-specific knowledge to improve the performance on rare labels.

\section*{Limitations}
Our proposed general knowledge injection framework, while offering an effective approach for the injection of knowledge to improve ICD coding performance, has notable limitations. First, it focuses on the ICD-9 code system, which, though widely used in prior research, is outdated compared to the more comprehensive ICD-10 system (e.g., over 70,000 diagnosis codes in ICD-10-CM vs. ~14,000 in ICD-9). Future work should adapt our approach to ICD-10.
Second, our framework does not incorporate the Alphabetic Index, a key tool in ICD coding. Coders use the Alphabetic Index to map clinical terms to a set of candidates before assigning the final ICD codes, ensuring accurate ICD coding. Future work should also integrate the Alphabetic Index.

\section*{Ethics Statement}
We use the publicly available clinical dataset
MIMIC-III, which contains de-identified patient
information. We do not see any ethics issues here in
this paper.

\section*{Acknowledgement}
Supported by the National Natural Science Foundation of China under Grant 62271465, the Suzhou Basic Research Program under Grant SYG202338, and the China Postdoctoral Science Foundation under Grant 2024M763178.

\bibliography{custom}



\end{document}